\documentclass{tlp}
\pdfoutput=1
\usepackage{multirow}
\usepackage{booktabs}
\usepackage{comment}
\usepackage{graphicx}
\usepackage[fleqn]{amsmath}
\usepackage{amssymb}
\usepackage{helvet,times,courier}
\usepackage{tikz}
\usepackage{url}
\usepackage{bold-extra}
\usepackage{bm}
\usepackage{listings}
\usepackage[utf8]{inputenc}
\usepackage[small]{caption}
\usepackage[hidelinks]{hyperref}
\usepackage{subfigure}
\usepackage{algorithm}
\usepackage{algorithmic}
\usepackage{tikz}

\usetikzlibrary{external}
\usetikzlibrary{positioning}
\usetikzlibrary{calc}
\usetikzlibrary{arrows}
\usetikzlibrary{decorations.pathmorphing,decorations.markings}
\usetikzlibrary{shapes}
\usetikzlibrary{patterns}

\newtheorem{proposition}{Proposition}
\newtheorem{example}{Example}
\newtheorem{theorem}{Theorem}

\newcommand{\ttb}[1]{\texttt{\textbf{#1}}}
\newcommand{\rg}[3]{{#1}{#2}\ldots{#2}{#3}}
\newcommand{\set}[1]{\{#1\}}
\newcommand{\union}{\cup}
\newcommand{\pair}[2]{\langle{#1,#2}\rangle}
\newcommand{\tuple}[1]{\langle{#1}\rangle}
\newcommand{\sel}[2]{\{{#1}\mid{#2}\}}
\newcommand{\GLred}[2]{{#1}^{#2}}

\newcommand{\at}[1]{\mathtt{#1}}
\newcommand{\sig}[1]{\mathrm{At}(#1)}
\newcommand{\compl}[1]{\overline{#1}}
\newcommand{\If}{\mathbin{\ttb{:-}\,}}
\newcommand{\Not}{\ttb{not}\,\,}
\renewcommand{\And}{\ttb{,}\,}
\newcommand{\Sep}{\ttb{;}\,}
\newcommand{\Or}{\ttb{|}\,}
\newcommand{\End}{\ttb{.}}
\newcommand{\asgn}{\leftarrow}

\newcommand{\as}[1]{\mathrm{AS}(#1)}
\newcommand{\eval}[2]{\mathrm{eval}_{#1}(#2)}
\newcommand{\constrainteq}[2]{\mathrm{EQ}_{#1}(#2)}
\newcommand{\constraintgt}[2]{\mathrm{GT}_{#1}(#2)}

\newcommand{\NP}{\mathbf{NP}}
\newcommand{\CONP}{\mathbf{coNP}}
\newcommand{\DP}{\mathbf{P}}
\newcommand{\EP}[1]{\Sigma^{\mathrm{p}}_{#1}}
\newcommand{\FP}[1]{\mathrm{FP}^{#1}}

\DeclareMathOperator*{\argmax}{arg\,max}

\begin{document}

\title{Solution Enumeration by Optimality \linebreak[4]
in Answer Set Programming}

\author[J. Pajunen and T. Janhunen]
{JUKKA PAJUNEN${~}^{1}$ and TOMI JANHUNEN${~}^{2,1}$ \\
${}^{1)}$%
Department of Computer Science, Aalto University \\
P.O.Box 15400, FI-00076 AALTO, Finland \\
\email{jukka.m.pajunen@gmail.com} \\
~\\
${}^{2)}$%
Computing Sciences, Tampere University \\
FI-33014 TAMPERE UNIVERSITY, Finland \\
\email{tomi.janhunen@\{tuni.fi,aalto.fi\}} \\
}

\maketitle

\begin{abstract}
Given a combinatorial search problem, it may be highly useful to
enumerate its (all) solutions besides just finding one solution,
or showing that none exists. The same can be stated about optimal
solutions if an objective function is provided.
This work goes beyond the bare enumeration of optimal solutions and
addresses the computational task of solution enumeration by optimality
(SEO). This task is studied in the context of Answer Set Programming
(ASP) where (optimal) solutions of a problem are captured with the
answer sets of a logic program encoding the problem. Existing
answer-set solvers already support the enumeration of all (optimal)
answer sets.
However, in this work, we generalize the enumeration of optimal answer
sets beyond strictly optimal ones, giving rise to the idea of answer
set enumeration in the order of optimality (ASEO). This approach is
applicable up to the best $k$ answer sets or in an unlimited setting,
which amounts to a process of sorting answer sets based on the
objective function.
As the main contribution of this work, we present the first general
algorithms for the aforementioned tasks of answer set enumeration.
Moreover, we illustrate the potential use cases of ASEO.
First, we study how efficiently access to the next-best solutions can
be achieved in a number of optimization problems that have been
formalized and solved in ASP.
Second, we show that ASEO provides us with an effective sampling
technique for Bayesian networks.
This article is under consideration for acceptance in TPLP.
\end{abstract}


\section{Introduction}

In this paper, we address \emph{combinatorial problem solving} where
some solution components are combined to meet problem specific
requirements. Such problems are frequent in computer science and
computationally hard to solve since the number of solution candidates
is usually exponential in the length of a problem instance
\cite{combinatorialComplex}.
Besides finding solutions for an instance, it is also possible to seek
solutions that are \emph{optimal} as defined by an \emph{objective
  function} $f$ mapping solutions to numbers. In fact,
\emph{combinatorial optimization problems} described, e.g., by
Korte and Vygen \citeyear{KV06:book}
tend to be computationally harder to solve, since the proof for
optimality presumes the exclusion of yet better solutions.
However, optimal solutions are not necessarily unique, suggesting the
search and \emph{enumeration} of all optimal solutions, e.g., for the
sake of generating alternatives or better understanding the true
nature of the objective function $f$.
Besides enumeration, Murty \citeyear{Murty68:or} was interested in the
\emph{ranking} of solutions based on their \emph{cost}, taking
systematically into consideration sub-optimal solutions. While this
idea is general by nature, it is applied to a specific problem where
cost-optimal assignments are sought based on a cost matrix.
If, in addition, a bound $k$ on the number of solutions is introduced,
we arrive at the identification of the best $k$ solutions
as in the case of the shortest path problem studied by
Lawler \citeyear{Lawler72:mansci}.

In this work, we adopt the ideas discussed above and concentrate on
the enumeration of optimal solutions but also take recursively into
consideration sub-optimal but \emph{next-best} solutions according to
$f$.  This gives rise to a systematic procedure that we coin as
\emph{solution enumeration by optimality} (SEO) which can also be
understood as \emph{sorting} solutions by their values obtained from
the objective function. Since the underlying solution space is usually
worst-case exponential, it is immediate that SEO poses computational
challenges in general. For simplicity, we confine our attention to
problems and solving paradigms yielding \emph{finite} solution spaces
in this work.

Besides tailoring native search algorithms, one viable way to solve
combinatorial (optimization) problems is to describe their solutions
in terms of constraints and to use existing solver technology for the
search of actual solutions. To this end, well-known approaches are
\emph{(maximum) Boolean satisfiability} (SAT/MaxSAT) \cite{BHvMW09:handbook},
\emph{integer linear programming} (ILP),
\emph{constraint programming} (CP) \cite{RvBW06:handbook}, and
\emph{answer set programming} (ASP) \cite{BET11:cacm}
with \emph{optimization} \cite{SNS02:aij,answersetopt}.
While the number of solutions is generally unbounded in ILP, MaxSAT,
CP, and ASP yield finite solution spaces in their standard use cases,
hence enabling the exhaustive enumeration of all solutions in finite time.

As regards systematic solution enumeration, ASP offers somewhat better
premises, since a typical ASP encoding aims at a one-to-one
correspondence between solutions and answer sets. Gebser et
al.~\citeyear{ConflictDrivenEnumeration} present dedicated algorithms
for answer set enumeration (ASE) that are able to operate in
\emph{polynomial space} and to \emph{project} answer sets with respect
to a user-defined signature. Furthermore, in the presence of an
objective function, the enumeration of optimal answer sets (ASO) is
similarly feasible, once the respective optimum value of the objective
function has been determined.
In contrast, the enumeration features of contemporary SAT/MaxSAT
solvers are quite limited although the algorithmic ideas are
described by Gebser et al.~\citeyear{GKS09:cpaior}.
For this reason, we concentrate on ASP solvers that natively support
\emph{solution enumeration} (SE) and
the \emph{enumeration} of \emph{optimal solutions} (SO)
as discussed above, but not SEO in any systematic way.

Our main goal is to combine both answer set enumeration (ASE) and the
enumeration of optimal answer sets (ASO) as the task of enumerating
answer sets in the order of optimality as determined by an objective
function $f$.
In ASP, such functions are usually
\emph{pseudo-Boolean expressions}
that assign integer weights to literals. The outcome, i.e.,
\emph{answer set enumeration by optimality} (ASEO)
realizes SEO under the view that answer sets bijectively capture the
solutions of a problem being solved. In the sequel, we will explore
ways to extend ASO procedures such that after enumerating all optimal
solutions the search is continued to enumerate next-best solutions,
and so on. Basically, this is achieved by performing ASO recursively
but a careless implementation of this strategy may jeopardize the
desired (polynomial) consumption of space.

When it comes to the enumeration of answer sets $S$ as guided by an
objective function $f(S)$, we will explore two mainstream
approaches. The first aims at enumerating all answer sets unless the
enumeration procedure is interrupted by the user. The second presumes
a parameter $k$ that gives the number of answer sets to be enumerated
and, thereafter, the best $k$ answer sets $S$ are sought given $f(S)$.
The latter approach opens up new possibilities for organizing ASEO
when $k$ is small enough for storing intermediate solutions in
memory. To realize these two approaches, we will deploy existing
enumeration and optimization algorithms of Gebser et
al.~\citeyear{ConflictDrivenEnumeration}
and their implementations in the \emph{Clingo} system.
The main contributions of our work are:
\begin{enumerate}
\item The concept of \emph{answer set enumeration by optimality}
  (ASEO) as a process where answer sets $S$ are produced in an order
  determined by an objective function $f(S)$.
\item Relating ASE, ASO, and ASEO from
  the perspective of computational complexity.
\item The development of basic algorithms that implement ASEO either
  in limited settings (the best $k$ answer sets) or without limitation
  (sorting answer sets subject to $f$).
\item An experimental evaluation of algorithms using
  ASP encodings of optimization problems.
\item Demonstrating the potential of ASEO in
  \emph{sampling} guided by objective functions $f$.
\end{enumerate}
It should be stressed that the main ideas of the paper can be
generalized for other problem solving paradigms supporting
optimization as long as the solution spaces are finite.

This paper is organized as follows. First, we briefly recall
some concepts of answer set programming (ASP) and optimization
in Section~\ref{section:preliminaries},
and assess the computational complexity of solution
enumeration by optimality (SEO) in the context of ASP (ASEO)
in Section \ref{section:complexity}.
Given these premises, we present our mixed algorithms for ASEO
in Section~\ref{section:methods}.
The efficiency of the novel ASEO algorithms is then studied in
Section~\ref{section:exproblems} by using optimization problems from
ASP competitions as benchmarks.
Yet further application is established in
Section~\ref{section:sampling}: it is shown and experimentally
verified that our ASEO algorithms provide a potentially effective
method for \emph{sampling} and \emph{approximate inference} on
Bayesian networks.
Section~\ref{section:discussion-and-conclusion} concludes the paper.


\section{Preliminaries}
\label{section:preliminaries}

The basic syntax of \emph{answer set programming} (ASP)
\cite{BET11:cacm} is based on \emph{rules}
\begin{equation}\label{eq:rule}
\rg{\at{h}_1}{\Or}{\at{h}_l}
\If\rg{\at{b}_1}{\And}{\at{b}_n}\And\rg{\Not\at{c}_1}{\And}{\Not\at{c}_m}\End
\end{equation}
where $\at{h}_i$:s, $\at{b}_i$:s, and $\at{c}_i$:s are \emph{atoms}.
A \emph{constraint} is a rule (\ref{eq:rule}) with an empty head ($l=0$).
A \emph{logic program} $P$ is a finite set of rules (\ref{eq:rule}).
The program $P$ is called \emph{normal} if $l\leq 1$ holds for every
rule (\ref{eq:rule}) and \emph{disjunctive}, otherwise. Given a set of
atoms $A$, the \emph{reduct} $\GLred{P}{A}$ of $P$ with respect to
$A$ contains a positive rule
$\rg{\at{h}_1}{\Or}{\at{h}_l}\If\rg{\at{b}_1}{\And}{\at{b}_n}$
for each rule (\ref{eq:rule}) of $P$ such that $l>0$ and no
negated atom $\at{c}_i$ is in $A$.
Given a logic program $P$, let $\sig{P}$ be the \emph{signature} of
$P$, i.e., the set of atoms that occur in $P$.
An \emph{answer set} $A\subseteq\sig{P}$ of $P$ (i) violates no
constraint of $P$ and (ii) is a subset-minimal set \emph{closed} under
$\GLred{P}{A}$, i.e., if $\rg{\at{b}_1\in A}{,}{\at{b}_n\in A}$ for
some rule
$\rg{\at{h}_1}{\Or}{\at{h}_l}\If\rg{\at{b}_1}{\And}{\at{b}_n}$
of $\GLred{P}{A}$, then at least one $\at{h}_i$ is in $A$.
Each program $P$ induces set of answer sets, denoted $\as{P}$,
usually capturing the solutions of a search problem encoded by $P$.

The syntax of ASP has been generalized in various ways as proposed,
e.g., by Simons et al.~\citeyear{SNS02:aij}, but many extensions can
be translated back into rules of the form (\ref{eq:rule}) using
transformations of Bomanson et al.~\citeyear{BGJ14:jelia} and Alviano
et al.~\citeyear{AFG15:tplp}.
To cater for optimization problems within ASP, it is possible to
introduce \emph{objective functions} to identify \emph{optimal} answer
sets. For the purposes of this work, an objective function $f(A)$ is
defined as a pseudo-Boolean expression
\begin{equation}\label{eq:objective-function}
\rg{w_1\times l_1~}{+}{~w_n\times l_n}=\sum_{i=1}^n (w_i\times l_i)
\end{equation}
where each $w_i$ is a non-negative integer \emph{weight} and each
$l_i$ is a \emph{literal}, i.e., either an atom `$\at{a}$' or its
negation `$\Not\at{a}$'. If desired, negative weights can also be
tolerated but typically reduced back to non-negative ones.
Given an answer set $A\in\as{P}$, the value $f(A)$ of
(\ref{eq:objective-function}) at $A$ is
\begin{equation}\label{eq:eval-objective}
\eval{A}{\sum_{i=1}^n (w_i\times l_i)}
=\sum\sel{w_i}{1\leq i\leq n,~A\models l_i}
\end{equation}
where $\models$ denotes the (standard) \emph{satisfaction} of a
literal $l_i$ in $A$: (i) $A\models\at{a}$ iff $\at{a}\in A$, and
(ii) $A\models\Not\at{a}$ iff $\at{a}\not\in A$.
An answer set $A\in\as{P}$ is \emph{optimal} (in the sense of
minimization) if $f(A')<f(A)$ for no $A'\in\as{P}$. Maximization
can also be supported by transforming $f$ into $f'$ suitable for
minimization \cite{SNS02:aij}.
Certain applications call for several \emph{objective functions} $f_i$
(\ref{eq:objective-function}) indexed by a \emph{priority level}
$1\leq i\leq p$. The resulting values (\ref{eq:eval-objective}) are
interpreted \emph{lexicographically}, i.e., $f_1(A)$ is the most
important, then $f_2(A)$, etc. So, given an answer set (candidate)
$A$, its (prioritized) objective value $f(A)$ is essentially a tuple
$\tuple{\rg{f_1(A)}{,}{f_p(A)}}$.


\section{Complexity Landscape}
\label{section:complexity}

In the following, we assume some familiarity with basic notions of
computational complexity; see, e.g., the book of Papadimitriou
\citeyear{Papadimitriou94:book} for an account.
As regards \emph{decision problems}, checking the existence of an
answer set for a normal (resp.~disjunctive) ground logic program given
as input is $\NP$-complete (resp.~$\EP{2}$-complete) as respectively
shown by Marek and Truszczy{\'n}ski \citeyear{MT91:jacm} and Eiter and
Gottlob \citeyear{EG95:amai}.
However, the task of computing an \emph{optimal} answer set $A$ for a
\emph{normal} logic program subject to an objective function $f$ forms
an $\FP{\NP}$-complete \emph{function problem} as established by
Simons et al.~\citeyear{SNS02:aij}. This means that only polynomially
many calls to an NP-oracle are required in the worst case.
Indeed, the last call is typically the most demanding one since it
excludes the existence of yet better answer sets. Once an optimal
answer set $A$ has been determined, the respective optimum value
$f(A)$ is naturally also known.

Let us then address the computational complexity of enumerating
solutions in an order determined by an objective function $f$. Our
first result concerns solution enumeration in general, i.e., the tasks
SE, SO, and SEO, identified earlier. It is assumed that a search
problem instance $P$ has some finite representation, ultimately
encoded as a string, based on which solutions $S$ can be sought and
enumerated, and the value $f(S)$ of the objective function $f$ is
computable.

\begin{proposition}\label{prop:seo-in-general}
Given a search problem instance $P$ and an objective function $f$,
(i)
the task SE is (constant-time) reducible to SO and SEO and
(ii)
the task SEO is not reducible to SE nor to SO in general.
\end{proposition}

\begin{proof}
(i)
The instance $P$ can be reduced to an instance $\pair{P}{f}$ of SO and SEO
by setting a constant value $f(S)=c$ for every solution $S$ of $P$. Then,
both SO and SEO implement SE since the order of solutions is irrelevant and
all solutions get enumerated in the end.

(ii)
Suppose that the instance $\pair{P}{f}$ of SEO were reducible to an
instance $P'$ of SE without an objective function. To enumerate
the solutions of $P$, the instance $P'$ must possess the same set
of solutions. However, since there is no objective function for $P'$
under SE, these solutions can be enumerated in any order, not
necessarily compatible with $f$.

Quite similarly, suppose that the instance $\pair{P}{f}$ of SEO is
reducible to an instance $\pair{P'}{f'}$ of SO. Since the latter
must yield all solutions to $P$, the instance $P'$ must have
the same solutions as $P$ and, in addition, each solution $S$ to $P'$
must be optimal given $f'$, i.e., $f'(S)$ is constant.  Thus, the
(optimal) solutions to $P'$ can be enumerated in any order, not
necessarily compatible with $f$.

To conclude, reductions are not feasible
in general, regardless of computational resources.
\end{proof}

On the other hand, if only the enumeration of the best $k$ solutions
is of interest, the respective enumeration task may become easier than
SEO: fewer solutions have to be enumerated. For sufficiently small
values of $k$, the task may be even easier than SO but harder than
SO if $k$ exceeds the number of best solutions. However, the number of
optimal solutions is not known in advance, which makes SEO an
attractive option for the user. Typical answer set solvers perform
SO with $k=1$ by default when optimizing answer sets.
Next we concentrate on the enumeration of answer sets only and gather
evidence that ASEO can be more demanding than ASE or ASO in
general. One way to implement ASEO is to use ASO repeatedly as a
subroutine and to exclude the solutions already enumerated by a
constraint. It is established below that the intermediate instances of
ASO remain equally hard enumeration problems in the worst case.

\begin{theorem}\label{theorem:residue-complexity}
Given a normal logic program $P$ subject to an objective function
$f$ of the form (\ref{eq:objective-function}) and a non-negative
integer lower bound $b$ for $f$, the task of computing the next-best
answer set $A$ subject to $f(A)\geq b$ is $\FP{\NP}$-complete.
\end{theorem}

\begin{proof}
The bound $b$ and the objective function $f(A)$ for a candidate
answer set $A$ can be incorporated by using a single \emph{weight rule}
$r$ \cite{SNS02:aij} that effectively expresses $f(A)\geq b$ and
by translating that rule into normal rules \cite{BGJ14:jelia}.
The translation $N(r)$ is feasible in polynomial time and $P\union
N(r)$ subject to $f$ is an (unbounded) instance of answer-set
optimization and, thus, a member in $\FP{\NP}$ as shown by Simons et
al.~\citeyear{SNS02:aij}.
Hardness for $\FP{\NP}$ is implied by the bound $b=0$
and the original (unbounded) complexity result
of Simons et al.~\citeyear{SNS02:aij}.
\end{proof}

Proposition \ref{prop:seo-in-general} does not cover the reducibility
of an instance $\pair{P}{f}$ of ASO with respect to ASE and
ASEO. Interestingly, if the objective value $f(A)=b$ for a best answer
set $A$ were known, reductions would become feasible.  By expressing
the condition $f(A)\leq b$ with a weight rule $r$ rewritten as a set
$N(r)$ of normal rules
(cf.~the proof of Theorem \ref{theorem:residue-complexity}),
the instance $P\union N(r)$ of ASE, or the instance
$\pair{P\union N(r)}{f}$ of ASEO, would capture the original ASO problem.
The key observation is that such a bound $b$ on $f$ cannot be
determined in general in polynomial time.

\begin{proposition}\label{prop:ASO-not-reducible}
The problem ASO is not polytime reducible to ASE nor
to ASEO unless $\DP=\CONP$.
\end{proposition}

\begin{proof}
Let $P$ be a normal program forming an instance of deciding if
$\as{P}=\emptyset$ or not. In general, this problem was shown
$\CONP$-complete by Marek and Truszczy{\'n}ski \citeyear{MT91:jacm}.
Let us construct a program $Q$ by
introducing new atoms $\at{a}$ and $\at{b}$ as well as rules
``$\at{a}\If\Not\at{b}\End$'', ``$\at{b}\If\Not\at{a}\End$'',
and the rules of $P$ conditioned by the atom $\at{b}$. Then
$\set{\at{a}}\in\as{Q}$ and, if $A\in\as{P}$, then
$A\union\set{\at{b}}\in\as{Q}$. By setting an objective function
$f(A)=2\times\at{a}+1\times\at{b}$, the answer set $\set{\at{a}}$ is
optimal only if answer sets of the latter kind do not exist.
Towards our goal, let us assume that $R$ is the polytime
reduction of $\pair{Q}{f}$ into an ASE problem. We can check in
polynomial time if $\set{\at{a}}\in\as{R}$ (i.e., whether it should be
enumerated by ASE), iff $\as{P}=\emptyset$.

Similarly, let $\pair{R}{g}$ be a polytime reduction of
$\pair{Q}{f}$ into an ASEO problem. Again, we can test if
$\set{a}\in\as{R}$ in polynomial time, giving rise to two cases.
(i)
If $\set{\at{a}}\not\in\as{R}$, then $Q$ has answer sets of the form
$A\union\set{\at{b}}$, all to be enumerated by ASEO and equally
admitted by $g$, i.e., $f(A\union\set{\at{b}})=b$ is
constant. Thus $\as{P}\neq\emptyset$.
(ii)
If $\set{\at{a}}\in\as{R}$, two subcases arise:
\begin{enumerate}
\item If $g(\set{\at{a}})$ is not minimal, then $R$ must
      have an answer set $B\neq\set{\at{a}}$ such that
      $g(B)<g(\set{\at{a}})$. Without loss of generality,
      assume that $g(B)$ is minimal. Then $B$ is to be enumerated by
      ASEO but also $\set{a}$, since ASEO will enumerate the
      inferior answer sets as well. Then both $B$ and $\set{a}$ should
      enumerated by ASO on the input $\pair{Q}{f}$, a contradiction.
\item Thus $g(\set{\at{a}})$ is minimal and $\set{\at{a}}$ is to be
      enumerated by ASEO. Then it must be enumerated by ASO
      given $\pair{Q}{f}$ excluding all other answer sets.
      Thus $\as{P}=\emptyset$.
\end{enumerate}
Thus we obtained a (hypothetical) polynomial-time test whether
$\as{P}=\emptyset$.
\end{proof}

\begin{proposition}\label{prop:exp-effects}
In the worst case, a (normal) logic program $P$ subject to an objective
function $f$ may possess exponentially many answer sets $A$ with
different objective values $f(A)$.
\end{proposition}

\begin{proof}
For an illustration, let us consider a concrete normal program $P_n$
based on atoms $\rg{\at{a}_1}{,}{\at{a}_n}$ and $\rg{\at{b}_1}{,}{\at{b}_n}$
parameterized by $n$. The intended objective function
$f(\rg{\at{a}_1}{,}{\at{a}_n})=\sum_{i=1}^n2^{i-1}\times\at{a}_i$.
Thus assignments to $\rg{\at{a}_1}{,}{\at{a}_n}$ encode
distinct values in the range $\rg{0}{,}{2^n-1}$ in binary.
Using \emph{complementary} atoms $\compl{\at{a}}$ for the atoms
$\at{a}$ involved, the rules of $P_n$ are for all $i=\rg{1}{,}{n}$:
\begin{center}
\begin{tabular}{llll}
$\at{a}_i\If\Not\compl{\at{a}}_i\End$
& $\compl{\at{a}}_i\If\Not\at{a}_i\End$
& $\at{b}_i\If\Not\compl{\at{b}}_i\End$
& $\compl{\at{b}}_i\If\Not\at{b}_i\End$ \\
$\at{lt}_i\If\compl{\at{a}}_i\And\at{b}_i\End$
& \multicolumn{2}{l}{%
$\at{lt}_i\If\at{a}_i\And\at{b}_i\And\at{lt}_{(i+1)}\And(i<n)\End$}
&
$\at{lt}_i\If\compl{\at{a}}_i\And\compl{\at{b}}_i\And
             \at{lt}_{(i+1)}\And(i<n)\End$ \\
$\If\at{b}_1\And\at{lt}_2\End$ &
$\If\compl{\at{b}}_1\And\Not\at{lt}_2\End$ \\
\end{tabular}
\end{center}
The rules of $P_n$ serve the following purposes.
The rules of the first line simply choose truth values for
$\at{a}_i$:s and $\at{b}_i$:s. The rules in the second line check
whether the value represented by $\at{a}_i$:s is \emph{lower than} the
value represented by $\at{b}_i$:s such that $\at{lt}_i$ takes bits
$\rg{i}{,}{n}$ into account. Regarding constraints in the third line,
they accept only one half of possible values for $\at{b}_i$:s given
some fixed values of $\at{a}_i$:s. Thus $P_n$ has $2^{n-1}$ answer
sets $A$ corresponding to each (fixed) assignment to the atoms
$\rg{\at{a}_1}{,}{\at{a}_n}$ and, in total, $2^{n}\times
2^{n-1}=2^{2n-1}$ answer sets.

Given an answer set $A$ of $P_n$, the objective value $f(A)$ is
$f(\rg{\at{a}_1}{,}{\at{a}_n})=\sum_{i=1}^n2^{i-1}\times\at{a}_i$
based on the truth values of $\rg{\at{a_1}}{,}{\at{a}_n}$ in $A$.
Given $P_n$ and $f$, any ASEO algorithm must enumerate all answer sets
$A$ with value $f(A)=0$ first, all answer sets with value $f(A)=1$
next, and so on, until (finally) all answer sets with value
$f(A)=2^{n}-1$ get enumerated.
Each subtask in this sequence can be viewed as an ordinary ASO task
yielding $2^{n-1}$ assignments for $\at{b}_i$:s and those assignments
are different for each assignment of $\at{a}_i$:s.  This is to prevent
the learning of nogoods for systematically excluding certain values of
$\at{b}_i$:s.
In this way, $P_n$ maximally exercises the underlying enumeration
procedure, although finding a single answer set is yet easy.
\end{proof}


\section{Algorithms for Answer Set Enumeration by Optimality}
\label{section:methods}

Propositions \ref{prop:seo-in-general}--\ref{prop:exp-effects}
support the view that ASEO can be strictly more demanding task than ASE
and ASO. This, however, does not prevent us from using
ASE and ASO as subroutines when realizing ASEO in practice.
Therefore, we implement our ASEO algorithms on top of a
state-of-the-art ASP solver
\emph{Clingo}~\cite{clingoUG}
treated as a \emph{black box} for running ASO and ASE subtasks for
systematic answer set enumeration. \emph{Clingo} implements an
efficient search for answer sets, as witnessed by ASP competitions
(see, e.g., the report of Calimeri et al.~\citeyear{fiftAspComp}), deploying
\emph{conflict-driven nogood learning}
(CDNL) algorithms devised by Gebser et al.~\citeyear{ConflictDrivenSolving}.
However, sub-optimal answer sets cannot be readily enumerated, thus
motivating the goals of our work. A way to implement ASEO is to use
\emph{Clingo}'s Python API that extends the functionality of the
underlying solver. While the API enables quite straightforward
implementation of ASEO algorithms detailed in the sequel, it excludes
certain options available for native implementations.

We call our first approach to ASEO the \emph{naive algorithm}. The
algorithm simply enumerates all \emph{answer sets} one by one with the
help of \emph{Clingo} API during which pairs $\pair{A}{f(A)}$ of
answer sets $A$ and the respective objective values
$f(A)=\tuple{\rg{f_1(A)}{,}{f_p(A)}}$
for priority levels are recorded for an input program $P$. Once the
ordinary ASE task is completed, the pairs recorded are \emph{sorted}
lexicographically based on their objective values. Then the output
consists of all answer sets of the program in the sorted order.
While our naive algorithm for ASEO is easy to implement, its
obvious weakness is its consumption of space. However, it may perform
surprisingly well in cases where the number of answer sets is
low. Therefore, it makes a reasonable baseline algorithm for our
purposes as well as a basis for comparisons against more sophisticated
algorithms.

\begin{algorithm}[tb]
\caption{Weight Enumeration}
\label{alg:enum-weight}
\begin{tabular}{l l}
\textbf{Input}: & Program: $P$, Integer: $k$ \\
& Objective functions:
$(1, (l^1_1, w^1_1),\dots,(l^1_{n_1}, w^1_{n_1})),\dots,(p, (l^p_1, w^p_1),\dots,(l^p_{n_p}, w^p_{n_p})$ \\
\textbf{Output}: & Sorted answer sets: $S$ \\
\end{tabular}
\begin{algorithmic}[1]
\STATE $S \asgn \{\}$; ~ $p^{\prime} \asgn p$; ~ $P^\prime \asgn P$
\WHILE{$p^{\prime}\geq 1$} \label{alg:weight:loop}
\STATE $A$ $\asgn$ ASP-Optimize($P^{\prime}$)
\IF{UNSAT($P^\prime$) and $p^{\prime}>1$} \label{alg:weight:rel1}
\IF{$|S| = 0$} \label{alg:weight:incons1}
\STATE \textbf{return} $S$
\ENDIF \label{alg:weight:incons2}
\STATE $p^{\prime} \asgn p^{\prime}-1$; ~ $P^\prime \asgn P$
\FOR{$i$ from $1$ to $p^{\prime}-1$}
\STATE $P^\prime \asgn P^\prime \cup \set{\constrainteq{i}{C[i]}}$
\ENDFOR
\STATE $P^\prime \asgn P^\prime \cup \set{\constraintgt{l}{C[p^{\prime}]}}$ \label{alg:weight:rel2}
\ELSE
\STATE $P' \asgn P$; ~ $C \asgn [~]$ \label{alg:weight:cprog1}
\FOR{$i$ from $1$ to $p$}
\STATE $C[i] \asgn \eval{A}{\sum_{j=1}^{n_i} (w_j^i \times l_j^i)}$; ~
$P' \asgn P' \cup \set{\constrainteq{i}{C[i]}}$
\ENDFOR
\FOR{$A^\prime \asgn$ ASP-Solve($P^\prime$)} \label{alg:weight:optN}
\STATE $S \asgn S \cup \set{A^\prime}$
\IF{$|S| = k$}
\STATE \textbf{return} $S$
\ENDIF
\ENDFOR \label{alg:weight:cprog2}
\STATE $p^{\prime} \asgn p$; ~ $P^\prime \asgn P$  \label{alg:weight:const1}
\FOR{$i$ from $1$ to $p^{\prime}-1$}
\STATE $P^\prime \asgn P^\prime \cup \set{\constrainteq{i}{C[i]}}$
\ENDFOR
\STATE $P^\prime \asgn P^\prime \cup \set{\constraintgt{l}{C[p^{\prime}]}}$
\ENDIF  \label{alg:weight:const2}
\ENDWHILE
\STATE \textbf{return} $S$
\end{algorithmic}
\end{algorithm}

In contrast with the naive algorithm, the ASEO task can be
alternatively implemented without enumerating all answer sets first.
Such a procedure is obtained by first (i) performing ASO followed by
(ii) the introduction of constraints that exclude solutions enumerated
so far. Finally, (iii) the ASO task is performed on increasingly
constrained problem instances until the desired number of \emph{answer
  sets} have been found. The constraints have to be introduced with
extra care so that the (polynomial) space consumption of the
underlying ASEO task is not jeopardized.
Algorithm~\ref{alg:enum-weight} is such a constraining algorithm that
augments the input program $P$ between subsequent enumeration
tasks. Once all \emph{currently optimal} answer sets have been
enumerated, it appends constraints that make subsequent answer sets
\emph{strictly worse} than those enumerated so far.
We know by Theorem \ref{theorem:residue-complexity} that this does not
necessarily make the remaining computational task any easier, i.e.,
the subsequent computation of next-best solutions can still be
challenging.
For a cost
$C = \eval{A}{\sum_{j=1}^{n_i} (w_j^i \times l_j^i)}$
associated with an answer set $A$ and a priority level $1\leq i\leq
p$, a rule-based \emph{constraint} $\constraintgt{i}{C}$,
mimicking the ASP core standard \cite{aspcore2}, is:
\begin{equation}\label{eq:methods:weight_constraint}
\If ~ \ttb{\#sum\{}\rg{w^{i}_1 ~ \ttb{:} ~ l^{i}_1}{\Sep}{w^{i}_{n_i} ~ \ttb{:} ~ l^{i}_{n_i}} \ttb{\}} ~ \ttb{<=} ~ C \End
\end{equation}
The objective value at the priority \emph{level} $1\leq i\leq p$
is fixed with a similar constraint $\constrainteq{i}{C}$:
\begin{equation}\label{eq:methods:eq_weight_constraint}
\If \ttb{\#sum\{} \rg{w^{i}_1 ~ \ttb{:} ~ l^{i}_1}{\Sep}{w^{i}_{n_i} ~ \ttb{:} ~ l^{i}_{n_i}} \ttb{\}} ~ \ttb{!=} ~ C \End 
\end{equation}

The \emph{weight enumeration} algorithm starts in
Line~\ref{alg:weight:loop} by discovering an optimal \emph{answer set}
$A$ using \emph{Clingo}'s optimization algorithm. The call
ASP-Optimize$(P^{\prime})$ returns \emph{an optimal answer set},
if such an answer set $A$ exists. Lines
\ref{alg:weight:incons1}--\ref{alg:weight:incons2}
cover programs having no answer sets.
Lines \ref{alg:weight:cprog1}--\ref{alg:weight:cprog2}
calculate \emph{objective values} $\rg{f_1(A)}{,}{f_p(A)}$ for
priority levels and introduce constraints
(\ref{eq:methods:eq_weight_constraint})
to enumerate all equally good \emph{answer sets}. Enumeration is done
by \emph{Clingo}'s answer set enumeration algorithm designed by Gebser
et al.~\citeyear{ConflictDrivenEnumeration}: it performs
branch-and-bound search based on literals in the input program and
conflicts that occur during the enumeration. In Line
(\ref{alg:weight:optN}),
the call ASP-Solve$(P^{\prime})$ receives all
\emph{answer sets} of $P^{\prime}$, one at a time.
Once these get enumerated,
Lines \ref{alg:weight:const1}--\ref{alg:weight:const2}
deploy constraints (\ref{eq:methods:eq_weight_constraint}) for levels
$i=\rg{1}{}{p^{\prime}-1}$ and a constraint (\ref{eq:methods:weight_constraint})
for level $p^{\prime}$ to find next-best answer sets.
Lines \ref{alg:weight:rel1}--\ref{alg:weight:rel2} are used to
virtually relax all constraints in the augmented program $P^\prime$ so
that next-best answer sets from the following priority level can be
enumerated.

\begin{example}
Consider an optimization program $P$ with three answer sets
$A_1$, $A_2$, and $A_3$ with objective values
$f(A_1)=\tuple{1,4,1}$,
$f(A_2)=\tuple{1,4,7}$, and
$f(A_3)=\tuple{1,7,4}$.
Algorithm \ref{alg:enum-weight} starts by discovering the (only)
optimal answer set $A_1$ and constrains the answer sets $A$ of $P$ by
$f_1(A) = 1$, $f_2(A)=4$, and $f_3(A)>1$.
These make $A_2$ optimal and once found, the last is revised to
$f_3(A) > 7$. As a result, no answer sets exist, the constraints for
$f_2$ and $f_3$ are superseded by $f_2(A)>4$. When $A_3$ is found, the
constraints are revised to $f_1(A)=1$, $f_2(A)=7$, and
$f_3(A)>4$. From this point onward, no answer sets are met, the level
$p^{\prime}$ is decreased down to $1$, and Algorithm \ref{alg:enum-weight}
terminates.
\end{example}

Few observations about Algorithm \ref{alg:enum-weight} are worthwhile.
First, it is designed to find the best $k$ answer sets as determined
by the given objective functions $\rg{f_1}{,}{f_p}$. If the full ASEO
is rather desired, the parameter $k$ should be set to $\infty$,
enforcing the algorithm to loop over all answer sets in an order
compatible with $\rg{f_1}{,}{f_p}$.
Second, the algorithm uses an (ordered) set $S$ as an intermediate
storage for answer sets. Instead of storing found answer sets in $S$,
an option is to process them immediately, e.g., by printing them
for the user. This saves the space taken by $S$.

\begin{proposition}\label{prop:enum-weight-polynomial}
Algorithm~\ref{alg:enum-weight} can be implemented so that
it runs in space polynomial in $|\sig{P}|$.
\end{proposition}

\begin{proof}
The ASE and ASO tasks are feasible in polynomial space, see the
CNDL-ENUM-ASP algorithm of Gebser et
al.~\citeyear{ConflictDrivenEnumeration} for
details. Using this algorithm via the \emph{Clingo} API,
Algorithm~\ref{alg:enum-weight} is able to enumerate
optimal as well as sub-optimal answer sets.
Lines \ref{alg:weight:rel1} and \ref{alg:weight:rel2} therein
\emph{relax} constraints used to exclude optimal answer sets
enumerated during previous solver calls by forgetting them and by
adding tightened ones. Thus the size of $P$ does not increase
indefinitely and Algorithm~\ref{alg:enum-weight} operates in space
comparable to CNDL-ENUM-ASP regardless the value of $k$.
\end{proof}

Due to our black-box implementation, Algorithm~\ref{alg:enum-weight}
may suffer from restarts after enumerating all answer sets pertaining
to a particular objective value.
The underlying solver could benefit from previously learned conflicts
and there is even further potential expected from keeping track of
seen non-optimal answer sets that could be reused in subsequent
runs. However, such functionalities call for a native enumeration
algorithm and/or modifications to the underlying solver.

\begin{algorithm}[tb] 
\caption{Smart Enumeration}
\label{alg:enum-smart}
\begin{tabular}{@{} l l}
\textbf{Input}: & Program: $P$, Integer: $k$ \\
& Objective functions:
$(1, (l^1_1, w^1_1),\dots,(l^1_{n_1}, w^1_{n_1})),\dots,(p, (l^p_1, w^p_1),\dots,(l^p_{n_p}, w^p_{n_p})$ \\
\textbf{Output}: & Sorted answer sets: $S$ \\ 
\end{tabular}
\begin{algorithmic}[1]
\STATE $S \asgn []$; ~ $t \asgn \tuple{\infty,\ldots,\infty}$; 
       ~ $L \asgn \{\}$; ~ $A \asgn \{\}$
\FOR{$L$ $\asgn$ ASP-Enumerate($P$)} \label{alg:smart:startloop}
\STATE $C \asgn []$ 
\FOR{$i$ from $1$ to $p$}
\STATE $C[i] \asgn \eval{L}{\sum_{j=1}^{n_i} (w_j^i \times l_j^i)}$ 
\ENDFOR
\IF{$L$ is complete} 
\STATE $S\mathrm{.insert}(\pair{L^{+}}{C})$ \label{alg:smart:candidate1}
\IF{$|S| > k$}
\STATE $S\mathrm{.remove}(k + 1)$ 
\STATE $\pair{A}{t} \asgn S[k]$
\ENDIF \label{alg:smart:candidate2}
\ELSIF{$C > t$} \label{alg:smart:nogood1}
\STATE $P\asgn P\union\set{\If L\End}$
\ENDIF \label{alg:smart:nogood2}
\ENDFOR \label{alg:smart:loop2}
\STATE \textbf{return} $S$
\end{algorithmic}
\end{algorithm}

Algorithm~\ref{alg:enum-smart} realizes a window-based approach where
answer sets are computed using an ASE process while keeping the best
$k$ answer sets $\rg{A_1}{,}{A_k}$ in memory. In addition, the
objective value $f(A_k)$ of the $k$:th best answer set is used as a
\emph{threshold} $t$ to discard found (partial) answer sets that are
worse than $A_k$ and thus not contributing to the top $k$ answer sets.
Our implementation of Algorithm~\ref{alg:enum-smart} relies much on
\emph{Clingo} API as it needs to keep track of assigned literals as
well as to introduce new \emph{nogoods} for excluding answer sets.
As regards storing answer sets, the space requirement of Algorithm~2
is $k\times|\sig{P}|$. If the underlying ASE algorithm were completely
used a black box, the actual enumeration of answer sets is feasible in
space polynomial in $|\sig{P}|$, but for sufficiently large values of
$k$, this is dominated by $k\times|\sig{P}|$. However, since
Algorithm~2 adds nogoods explicitly and for a different purpose than
the underlying ASE algorithm, their number may become exponential in
$|\sig{P}|$ in contrast with Algorithm 1
(cf. Proposition \ref{prop:enum-weight-polynomial}).
This issue can be best approached in terms of either native algorithms
or an enhanced API that is able to keep track and manipulate nogoods
used for different purposes.

The main loop of \emph{smart enumeration} in Lines 
\ref{alg:smart:startloop}--\ref{alg:smart:loop2}
start with a call to ASP-Enumerate$(P)$ that is based on
the standard ASE algorithm of
Gebser et al.~\citeyear{ConflictDrivenEnumeration}.
The call provides a new \emph{partial assignment} $L$, i.e., a set of literals
for the input program $P$. For each iteration of the loop, $L$ is
extended by some chosen literal $l$ and further literals may be derived
using propagation. The underlying enumeration algorithm is also
responsible for backtracking if $L$ becomes inconsistent. Note that
since $L$ is partial, (\ref{eq:eval-objective}) is revised to
$\eval{L}{\sum_{i=1}^n (w_i\times l_i)}
 =\sum\set{w_i\mid 1\leq i\leq n, l_i\in L}$.
If $L$ is \emph{complete} (covers all of $\sig{P}$), Lines
\ref{alg:smart:candidate1}--\ref{alg:smart:candidate2}
append the respective answer set
$L^+=\sel{\at{a}\in L}{\at{a}\in\sig{P}}$
and the corresponding cost(s) $C$ to the list $S$ of best $k$
solutions so far. Moreover, the threshold value $t$ for excluding
future answer sets is updated. Line \ref{alg:smart:candidate1}
implements \emph{insertion sort} and places $\pair{L^+}{C}$ in the
list $S$ based on its cost(s) $C$.
However, if $L$ is not yet \emph{complete} and its cost vector
$f(L)$ is greater than the current threshold value $t$, the program
$P$ can be extended by a \emph{nogood} for excluding all supersets of
$L$ as done in Lines
\ref{alg:smart:nogood1}--\ref{alg:smart:nogood2}.
The algorithm terminates once the whole search space limited by the
nogoods is exhausted. 

\begin{example}
Consider a single objective function $f$ as defined by
$(l_1, 5)$,
$(l_2, 1)$,
$(l_3, 2)$,
$(l_4, 2)$,
$(l_5, 6)$
and all possible answer sets 
$A_1 = \{ l_1, l_2, l_3 \}$,
$A_2 = \{ l_1, l_3, l_5 \}$,
$A_3 = \{ l_2, l_3, l_5 \}$,
$A_4 = \{ l_1, l_2, l_4\}$,
$A_5 = \{ l_1, l_4, l_5 \}$
of a program $P$ for which the best $k=2$ answer sets are
sought. Suppose that $f(A_4)=8$ and $f(A_3)=9$ are the currently known
top $k=2$ answer sets which gives threshold $t=9$. If $L=\set{l_1}$ is
the current partial answer set and the search heuristic picks $l_5$ as
the next literal to branch on, we end up with $L^\prime=\set{l_1,l_5}$
with cost $f(L')=11$. Since $11 > t$, any extensions of $L^\prime$ are
eliminated by a nogood $\If l_1,l_5\End$ Thus the underlying ASP
solver will skip $A_2$ and $A_5$ altogether as it cannot construct
the corresponding partial assignments anymore.
\end{example}

Algorithms~\ref{alg:enum-weight} and \ref{alg:enum-smart} work quite
differently. For the program $P_n$ from the proof of Proposition
\ref{prop:exp-effects}, there exist exponentially many
\emph{answer sets} while finding and enumerating \emph{optimal values}
is easy.
Using Algorithm~\ref{alg:enum-weight} on this problem instance is fast
due to easy enumeration while Algorithm~\ref{alg:enum-smart} slows
down as it needs to find \emph{objective values} among an exponential
number of possibilities.


\section{Benchmarking: Optimization Problems}
\label{section:exproblems}

\begin{table}
\caption{Experimental results for ASEO algorithms}
\label{tab:results}
\begin{minipage}{\textwidth}
\begin{tabular}{ l c r r r r r r r }
\hline\hline
\multicolumn{4}{ c }{Instance} & \multicolumn{4}{ c }{Target Answer Sets $k$\footnote{Search up to $k$ Answer Sets}} & \\ 
\textbf{Problem} & \textbf{\#Instances} &  \textbf{Naive}\footnote{Average runtime, (number of timeouts) for Naive algorithm} & \textbf{Alg}\footnote{Which ASEO algorithm was used} & \textbf{$10$} & \textbf{$100$} & \textbf{$10^3$} & \textbf{$10^4$} & \textbf{\#TO}\footnote{Number of timeouts in instance} \\ 
\midrule
\multirow{2}{*}{BayesianNL} & \multirow{2}{*}{$30$} & \multirow{2}{*}{$1800s$ $(30)$} & Alg~1 & $\pmb{41s}$  & $\pmb{67s}$  & $\pmb{171s}$  & $\pmb{460s}$ & $\pmb{2}$ \\
& & & Alg~2 & $919s$  & $991s$  & $1083s$  & $1379s$ & $64$ \\
\noalign{\vspace {.5cm}}
\multirow{2}{*}{MarkovNL} & \multirow{2}{*}{$30$} & \multirow{2}{*}{$1800s$ $(30)$} & Alg~1 & $\pmb{185s}$  & $\pmb{311s}$  & $\pmb{757s}$  & $\pmb{1499s}$ & $\pmb{28}$ \\
& & & Alg~2 & $1800s$  & $1800s$  & $1800s$  & $1800s$ & $110$ \\
\noalign{\vspace {.5cm}}
\multirow{2}{*}{Supertree} & \multirow{2}{*}{$35$} & \multirow{2}{*}{$1099s$ $(10)$} & Alg~1 & $457s$  & $591s$  & $696s$  & $855s$ & $32$ \\
& & & Alg~2 & $\pmb{391s}$  & $\pmb{417s}$  & $\pmb{507s}$  & $\pmb{830s}$ & $\pmb{29}$ \\
\noalign{\vspace {.5cm}}
\multirow{2}{*}{Connected} & \multirow{2}{*}{$20$} & \multirow{2}{*}{$301s$ $(0)$} & Alg~1 & $\pmb{1613s}$  & $1713s$  & $1771s$  & $1800s$ & $73$ \\
& & & Alg~2 & $1685s$  & $\pmb{1684s}$  & $\pmb{1702s}$  & $\pmb{1760s}$ & $\pmb{72}$ \\
\noalign{\vspace {.5cm}}
\multirow{2}{*}{MaxSAT} & \multirow{2}{*}{$10$} & \multirow{2}{*}{$1800s$ $(10)$} & Alg~1 & $\pmb{21s}$  & $\pmb{721s}$  & $\pmb{902s}$  & $\pmb{919s}$ & $\pmb{14}$ \\
& & & Alg~2 & $1098s$  & $1368s$  & $1790s$  & $1800s$ & $32$ \\
\noalign{\vspace {.5cm}}
\multirow{2}{*}{FastFood} & \multirow{2}{*}{$20$} & \multirow{2}{*}{$212s$ $(1)$} & Alg~1 & $\pmb{1s}$  & $\pmb{1s}$  & $\pmb{7s}$  & $500s$ & $\pmb{0}$ \\
& & & Alg~2 & $10s$  & $9s$  & $31s$  & $\pmb{185s}$ & $1$ \\
\noalign{\vspace {.5cm}}
\multirow{2}{*}{Hamilton} & \multirow{2}{*}{$45$} & \multirow{2}{*}{$46s$ $(0)$} & Alg~1 & $\pmb{126s}$  & $350s$  & $797s$  & $1243s$ & $32$ \\
& & & Alg~2 & $187s$  & $\pmb{216s}$  & $\pmb{261s}$  & $\pmb{394s}$ & $\pmb{18}$ \\
\hline\hline
\end{tabular}
\vspace{-2\baselineskip}
\end{minipage}
\end{table}

The performance of our ASEO algorithms is evaluated on several
\emph{optimization problems} adopted from ASP competitions, as
reported by Calimeri et al.~\citeyear{fiftAspComp}.
The criterion for choosing problems was solvability by \emph{Clingo}
in a reasonable time.
For each benchmark problem, the difficulty of instances increases and
the numbers of answer sets are problem-specific. In the experiments, we
measure runtime subject to a timeout of $1\,800\,s$. All experiments
are conducted using an
\emph{Intel(R) Xeon(R)} CPU (\emph{E5-1650 v4 @ 3.60GHz}),
with 32GB RAM, and \emph{Ubuntu} OS.

Table~\ref{tab:results} collects the results obtained for optimization
problems. The rows of the table detail runs on different problems
while columns denote different runs of our algorithms. The second
column shows average time for the \emph{naive algorithm} as well as
number of timeouts in parentheses. The last column shows the
collective number of timeouts (\#TO) for an enumeration algorithm.  A
single cell reports the average runtimes for all instances of the
benchmark problem in question, and for Algorithms
\ref{alg:enum-weight} and \ref{alg:enum-smart}. For each target of
enumeration, the better algorithm is indicated in boldface.
The average runtimes suggest that the \emph{weight enumeration}
algorithm tends to perform better than \emph{smart enumeration}.  Even
when smart enumeration gives faster results, the runtime of weight
enumeration is not too far off.
Average runtimes obtained for \emph{BayesianNL} and \emph{MaxSAT}
highlight how much faster weight enumeration can be while
\emph{Hamilton} is the only benchmark with a reverse effect.
Yet it is important to note that for certain benchmarks,
namely \emph{Connected} and \emph{Hamilton},
the naive algorithm that enumerates all answer sets first
works faster than others.

\begin{figure}[tbp]
\begin{center}
\begin{tabular}{@{}c@{\hspace{0.02\linewidth}}c@{}}
\includegraphics[width=0.49\linewidth,trim={5pt 23pt 35pt 20pt},clip]{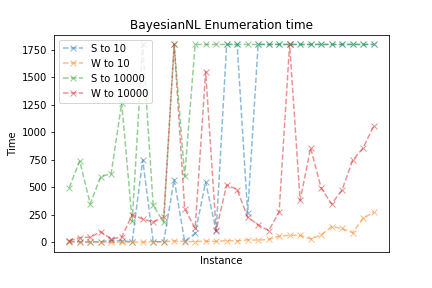}
&
\includegraphics[width=0.49\linewidth,trim={5pt 23pt 35pt 20pt},clip]{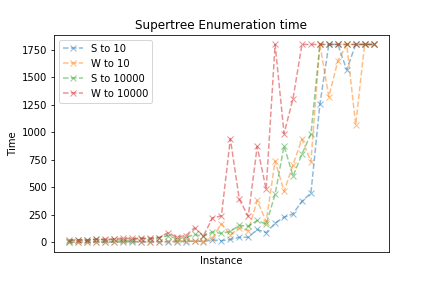}
\\
\begin{small}(a) Runtimes on \emph{BayesianNL}\end{small} &
\begin{small}(b) Runtimes on \emph{Supertree}\end{small}
\end{tabular}
\caption{Cactus plots for selected experiments \label{fig:results:all}}
\end{center}
\end{figure}

Fig.~\ref{fig:results:all} reflects how enumeration algorithms scale
differently. Fig.~\ref{fig:results:all}(a) reveals that
\emph{BayesianNL} is relatively easy for \emph{weight enumeration} (W).  The
switch from enumerating $10$ answer sets to $10^4$ answer sets seems to
increase the runtime of weight enumeration somewhat linearly.
Meanwhile, the performance of \emph{smart enumeration} (S) algorithm
degrades rabidly to the timeout.
In the same vein, we observe from Fig.~\ref{fig:results:all}(b)
that the two enumeration algorithms scale quite similarly in the
\emph{Supertree} benchmark problem. Their runtimes start to increase
simultaneously but the increase is more drastic for
weight enumeration than it is for smart enumeration,
and the latter algorithm runs always slightly faster than the former.
It is evident that the algorithms behave differently across the
instances and relative to each other. Algorithm~\ref{alg:enum-weight}
works considerably better than others when finding an optimal
objective value and enumerating optimal values is easy as, for
example, in the case of the logic program introduced in
the proof of Proposition \ref{prop:exp-effects}.
However, enumeration based algorithms work better when the number of
answer sets is reasonably low such that they can be enumerated in
comparable time.
Algorithm~\ref{alg:enum-smart} seems to perform especially well when
the objective values $\rg{f(A_1)}{\leq}{f(A_k)}$ for the best $k$
answer sets $\rg{A_1}{,}{A_k}$ vary a lot and it is easier to locate the
threshold value $t=f(A_k)$ for ruling out worse answer sets.

To summarize the experiments reported in this section, Algorithm
\ref{alg:enum-weight} (\emph{weight enumeration}) works reasonably
well for a majority of the problem instances. It seems to be able to
discover next-best answer sets with ease similar to the previous ones
and our best experimental results indicate linear increase in time
with respect to the desired number of answer sets $k$.
Meanwhile, Algorithm \ref{alg:enum-smart} (\emph{smart enumeration})
does not perform as well on average and it fails on problem instances
that are easy for weight enumeration. However, we are able to
demonstrate that there exist problems for which the smart enumeration
algorithm can be faster and thus beneficial.


\section{Practical Application: Bayesian Sampling}
\label{section:sampling}

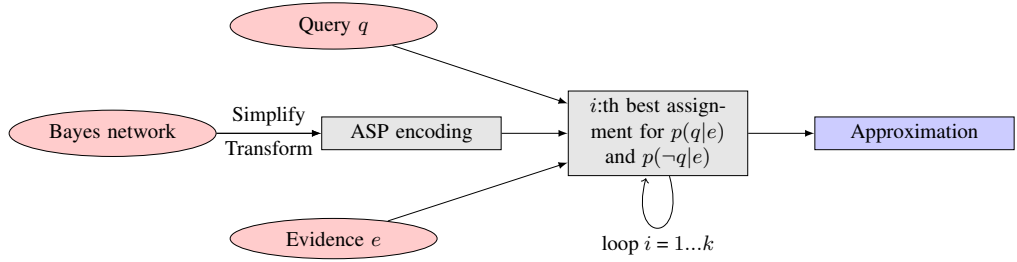
\begin{figure}
\centering
\scalebox{0.88}{\begin{tikzpicture}[
node distance=0.9cm and 1cm,
mynode/.style={draw,rectangle,fill=black!10,text width=2.5cm,align=center},
mycirc/.style={draw,fill=red!20,ellipse,text width=2cm,align=center},
asdnode/.style={draw,fill=blue!20,rectangle,text width=2.8cm,align=center}
]

\node[mycirc] (bn) {Bayes network};
\node[mynode, right=5em of bn] (tr) {ASP encoding};
\node[mycirc, above right=5em of bn] (q) {Query $q$};
\node[mycirc, below right=5em of bn] (e) {Evidence $e$};
\node[mynode, right=of tr] (MAP) {$i$:th best assignment for $p(q|e)$ and $p(\neg q | e)$};
\node[asdnode, right=of MAP] (app) {Approximation};

\path
(bn) edge [-latex,above] node {Simplify} (tr)
(bn) edge [-latex,below] node {Transform} (tr)
(tr) edge [-latex] (MAP)
(q) edge [-latex] (MAP)
(e) edge [-latex] (MAP)
(MAP) edge [loop below] node {loop $i$ = $1...k$} (MAP) 
(MAP) edge [-latex] (app);
\end{tikzpicture}}
\vspace{0.5\baselineskip}
\caption{ASEO-based approximation of probabilities}
\label{fig:procedure:map}
\end{figure}
  
\begin{table}[tbp]
\caption{Bayesian networks used in the experiments}
\label{tab:samp:networks}
\centering
\begin{tabular}{ c c c c c }  
\toprule
\multicolumn{4}{ c }{Network Features} & Algorithm Features \\
\textbf{Network} & \textbf{Nodes} & \textbf{Arcs} & \textbf{Avg.~Degree} & $k$ \\
\midrule
\noalign{\vspace {.1cm}}
\emph{Boe 92}    &  23            &  36           & 3.13 & $200$ \\ 
\noalign{\vspace {.1cm}}
\emph{Win-95}    &  76            & 112           & 2.95 & $500$ \\ 
\noalign{\vspace {.1cm}}
\emph{Andes}     & 223            & 338           & 3.03 & $2000$ \\
\noalign{\vspace {.1cm}}
\bottomrule
\end{tabular}
\end{table}

For a practical application of ASEO, let us consider
\emph{probabilistic inference} in the context of
\emph{Bayesian networks} (BNs)
with Boolean random variables and an approach
similar to \emph{weighted model counting} \cite{CD08:aij}.
The goal is to approximate queries by computing Maximum a
Posteriori (MAP) assignments using an ASP encoding of the respective
abduction problem devised by Beaver and Niemel{\"a} \citeyear{MAPtoASP}.
The optimal solutions of the problem satisfy
\begin{equation}\label{eqn:sampling:MAP}
c_{\mathrm{MAP}} = \argmax_C \left( p(C|e) \right),
\end{equation}
i.e., $c_{\mathrm{MAP}}$ is an assignment to \emph{random variables} $C$
in a network $N$ that makes the observed \emph{evidence} $e$ in $N$ most
probable. Given a (Boolean) \emph{query} variable $q$, the same ASP
encoding and the ASEO algorithms proposed in this work can be used to
enumerate assignments in a decreasing order of significance, i.e.,
starting from MAP assignments but continuing with next probable
assignments, and so on. This leads to a goal directed sampling
method as follows.
Given a query $q$ and some evidence $e$, we find out the $k$ most
probable assignments $\rg{c_1}{,}{c_k}$ (resp.~$\rg{d_1}{,}{d_k}$)
that satisfy $q$ (resp.~$\neg q$) and are compatible with $e$.
Thus, we obtain an estimate
\begin{equation}\label{eqn:sampling:approx}
P(q|e)\approx
\frac{\sum_{i=1}^kP(q,c_i \mid e)}{\sum_{i=1}^kP(q,c_i \mid e)+\sum_{i=1}^kP(\neg q,d_i \mid e)}.
\end{equation}

\begin{figure}[tbp]
\begin{center}
\begin{tabular}{@{}c@{\hspace{0.02\linewidth}}c@{}}
\includegraphics[width=0.49\linewidth,trim={60pt 40pt 50pt 50pt},clip]{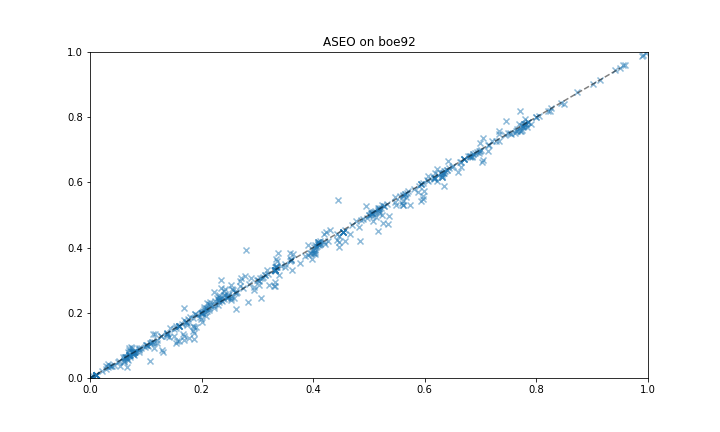}
&
\includegraphics[width=0.49\linewidth,trim={60pt 40pt 50pt 50pt},clip]{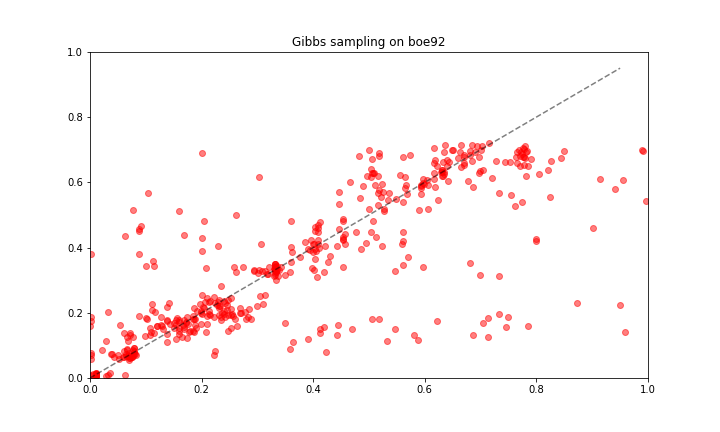}
\\
\begin{small}(a) ASEO on \emph{Boe 92}\end{small} &
\begin{small}(b) Gibbs on \emph{Boe 92}\end{small}
\\
\includegraphics[width=0.49\linewidth,trim={60pt 40pt 50pt 50pt},clip]{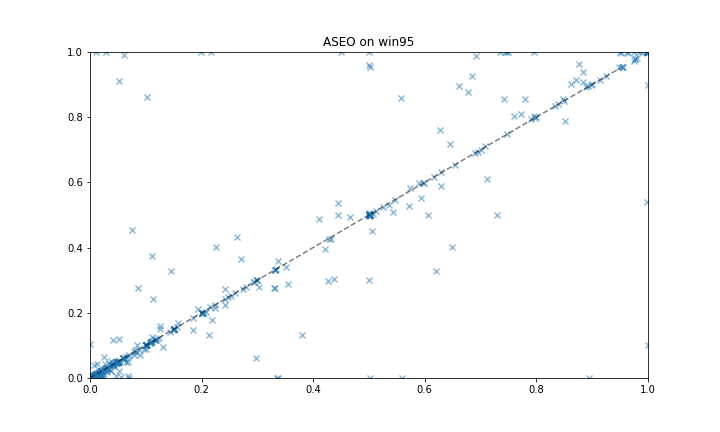}
&
\includegraphics[width=0.49\linewidth,trim={60pt 40pt 50pt 50pt},clip]{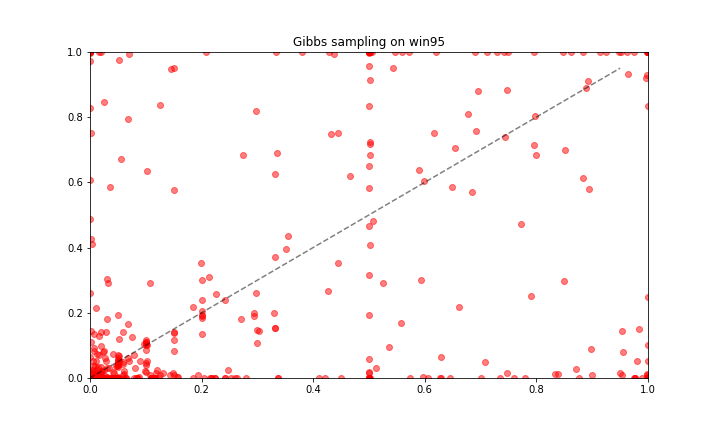}
\\
\begin{small}(c) ASEO on \emph{Win-95}\end{small} &
\begin{small}(d) Gibbs on \emph{Win-95}\end{small}
\\
\includegraphics[width=0.49\linewidth,trim={60pt 40pt 50pt 50pt},clip]{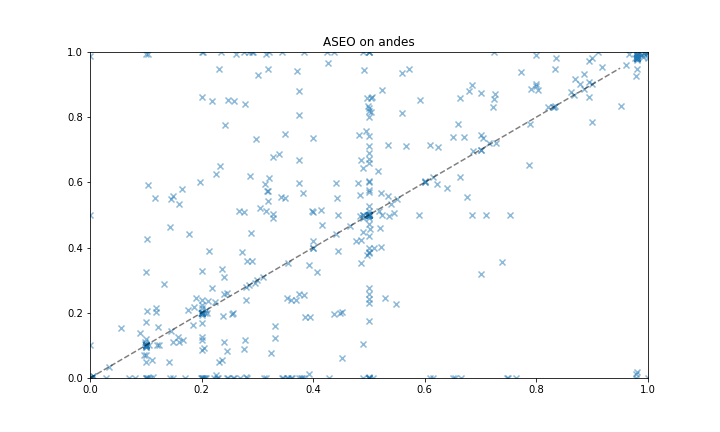}
&
\includegraphics[width=0.49\linewidth,trim={60pt 40pt 50pt 50pt},clip]{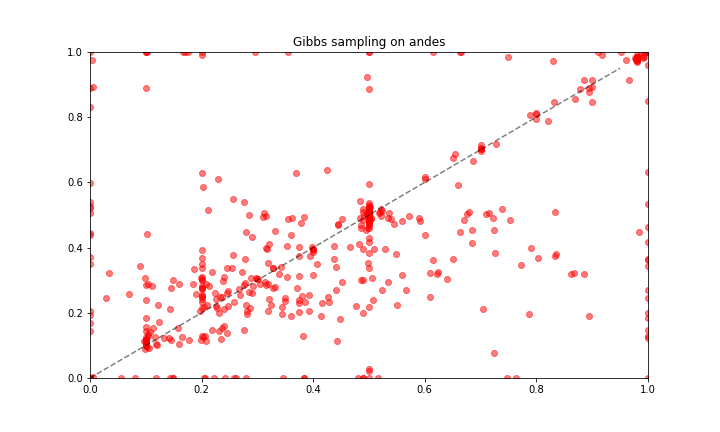}
\\
\begin{small}(e) ASEO on \emph{Andes}\end{small} &
\begin{small}(f) Gibbs on \emph{Andes}\end{small}
\end{tabular}
\caption{Results of approximation algorithms compared to true probabilities in the networks \label{fig:samp_results}}
\end{center}
\end{figure}

Figure~\ref{fig:procedure:map} illustrates the full procedure for
\emph{approximating} the query $q$ given the evidence $e$. First, for
the sake of fair comparison with other sampling methods, the network
is simplified using \emph{d-separation} following Geiger et
al.~\citeyear{GVP89}.
Second, the input network is transformed into an ASP encoding for
solving the respective MAP problem. Then, based on the given evidence
$e$, the query $q$, and its negation $\neg q$, up to $k$ most probable
assignments are computed in order to approximate the probability
$p(q|e)$ in the sense of (\ref{eqn:sampling:approx}).
Then, we evaluate our ASEO-based sampling method using few publicly
available BNs whose size varies from small to large (cf.~Table
\ref{tab:samp:networks}). The experiments are conducted by
\emph{randomly} picking a query variable and a set of evidence variables
that represent $1$ to $50\%$ of the variables in the network in
total. In what follows, we compare our sampling results to those
obtained with a well-known native algorithm, viz.~\emph{Gibbs sampler}
as provided in the \emph{pyAgrum} library of Gonzales et
al.~\citeyear{pyAGRUM}. In this implementation, the sampler stops by
determining the convergence of the approximation with $\epsilon=0.01$.

\begin{table}
\caption{Measured errors of approximations}
\label{tab:avgerror}
\begin{minipage}{\textwidth}
\begin{tabular}{ c c c c }
\hline\hline
\multicolumn{4}{l}{\emph{Average distance }} \\
\textbf{Algorithm} & \textbf{Boe 92} & \textbf{Win-95} & \textbf{Andes} \\ \midrule
\noalign{\vspace {.1cm}}
\emph{ASEO}              & $0.03$    & $0.04$   & $0.16$          \\
\noalign{\vspace {.1cm}}
\emph{Gibbs sampler}  & $0.07$    & $0.11$  & $0.15$           \\ 
\hline\hline
\multicolumn{4}{l}{\emph{Maximum error}} \\
\textbf{Algorithm} & \textbf{Boe 92} & \textbf{Win-95} & \textbf{Andes} \\ \midrule
\noalign{\vspace {.1cm}}
\emph{ASEO}              & $0.11$    & $0.98$   & $0.99$ \\ 
\noalign{\vspace {.1cm}}
\emph{Gibbs sampler}  & $0.81$    & $0.99$  & $0.99$ \\ 
\hline\hline
\end{tabular}
\end{minipage}
\end{table} 

In Fig.~\ref{fig:samp_results}, the sampling results for the input
networks are presented. To obtain the results we use
Algorithm~\ref{alg:enum-weight} to conduct ASEO where the choice of
the enumeration algorithm was based on better overall performance. In
each image, true probabilities are mapped on the $x$-axis while the
respective estimates are represented on the $y$-axis. Hence, the
closer to the diagonal points are, the better approximations have been
obtained. When looking at distributions, we observe that the
ASEO-based approximations tend to be better on smaller networks while
this is no longer obvious for the largest network \emph{Andes}. As far
as we can see, this is due to the fact that with larger networks there
exists increasing number of assignments with an equal objective value
(i.e., the same probability) and individual assignments having very
low probability values. As a consequence, we do not get a good
overview of the distribution of probabilities and the probabilities
are far from the expected true ones as for the \emph{Andes} network.

Table~\ref{tab:avgerror} presents further qualitative results over
approximations where \emph{distances} to true probabilities are
measured. The results of the table illustrate that ASEO-based sampling
can be competitive even with known methods such as Gibbs
sampler. However, when considering approximate reasoning task like
Bayesian sampling, one should remember that there are highly optimized
\emph{exact} methods, such as that of Madsen and Jensen
\citeyear{MJ99:aij}, to determine true probabilities fast.


\section{Discussion and Conclusion}
\label{section:discussion-and-conclusion}

In this work, we address solution enumeration in the context of
combinatorial optimization problems, the goal of which is to generate
all optimal solutions to a given problem instance. As a novelty, we
take also sub-optimal solutions into consideration and propose
algorithms that are able to recursively enumerate next-best solutions
until (i) all solutions or, alternatively, (ii) the best $k$ solutions
have been enumerated. Such a procedure realizes our concept of
\emph{solution enumeration by optimality} (SEO).
Besides screening the computational cost of SEO, we present dedicated
SEO algorithms geared toward answer set programming (ASP) where
solutions are captured with answer sets. The resulting ASEO algorithms
are implemented in Python \cite{Pajunen2020} using the API of the
\emph{Clingo} solver and are put to the test in experiments.
These algorithms generalize the ASO reasoning mode of \emph{Clingo}.
Moreover, we mention that the computation of $k$ best solutions was
considered in an experimental track of MaxSAT Evaluation run by
Bacchus et al.~\citeyear{MaxSAT},
but the solutions enumerated were supposed to assign different
values to literals involved in the objective function. Thus,
such an enumeration differs from SEO up to $k$ solutions.

The experimental evaluation comprises of two parts.
First, we evaluate our ASEO algorithms on a number of optimization
problems that have been used in ASP competitions. The results indicate
the feasibility of ASEO in practice, enabling the exploitation of
next-best solutions when solving optimization problems.
This is good news from the perspective of Theorem
\ref{theorem:residue-complexity} which indicates that the remaining
computational complexity does not necessarily decrease when excluding
answer sets already enumerated.
Second, we illustrate the potential of ASEO in approximating Bayesian
inference. There is some resemblance to the approach of Chavira and
Darwiche \citeyear{CD08:aij} based on weighted model counting. The
difference is that the objective function derived from conditional
probability tables guides the search toward the most relevant
assignments for query evaluation.
Ermon et al.~\citeyear{EGSS13} deploy \emph{optimization oracles} for similar
inference but in the presence of further constraints for uniformly
distributing samples over the space of assignments.
It is worth noting that Bayesian inference has been well studied and
there are further good alternatives to ASEO. Our main objective is to
demonstrate the new way of utilizing objective functions, effectively
SEO, as means to reach the most significant solutions first.

As discussed in Section \ref{section:methods}, our proof-of-concept
implementations of ASEO are still sub-optimal as they are essentially
based on successive calls to ASO/ASE algorithms via an API. This
prevents, e.g., the direct exploitation of nogoods learned during the
subsequent invocations of the underlying enumeration algorithm. This
observation suggests an obvious goal for future work: a native
implementation of an ASEO \emph{reasoning mode} in some
state-of-the-art answer-set solver.
To this end, we hope that our results make this goal attractive for
developers.


\paragraph{Acknowledgments.}

The second author has been partially supported by the Academy of
Finland projects ETAIROS (327352) and AI-ROT (335718).




\end{document}